\documentclass{llncs}

\usepackage[USenglish]{babel} 
\usepackage[T1]{fontenc}
\usepackage[ansinew]{inputenc}
\usepackage{amssymb,amsmath}
\usepackage{algorithm,algorithmic}
\usepackage{comment}
\usepackage{url}
\usepackage{cite}
\usepackage{graphicx}

\usepackage{tikz}
\usetikzlibrary[patterns, decorations.pathreplacing]

\usepackage[mathscr]{euscript}

\urldef{\mailsa}\path|giacomo.kahn@isima.fr|
\urldef{\mailsb}\path|contact@alexandrebazin.com|

\begin{document}

\begin{frontmatter}

\title{Average Size of Implicational Bases}
\titlerunning{Average size of implication base}
\author{Giacomo Kahn\inst{1}\and Alexandre Bazin\inst{2}}
\institute{LIMOS \& Universit\'e Clermont Auvergne, France \\
\and
Le2i - Laboratoire Electronique, Informatique et Image, France\\
\mailsa, \mailsb\\
}

\maketitle

\begin{abstract}
Implicational bases are objects of interest in formal concept analysis and its applications.
Unfortunately, even the smallest base, the Duquenne-Guigues base, has an exponential size in the worst case.
In this paper, we use results on the average number of minimal transversals in random hypergraphs to show that the base of proper premises is, on average, of quasi-polynomial size.
\end{abstract}
\begin{keywords}
Formal Concept Analysis, Implication Base, Average Case Analysis.
\end{keywords}
\end{frontmatter}

\section{Introduction}



Computing an implication base is a task that has been shown to be costly~\cite{DBLP:journals/dam/DistelS11}, due to their size and to the enumeration delay.
Even the smallest base (the Duquenne-Guigues base) is, in the worst case, exponential in the size of the relation~\cite{DBLP:journals/jucs/Kuznetsov04}.
While the extremal combinatorics of implicational bases is a well studied subject, up to now, the average case has not received a lot of attention.

\medskip

In this paper, we adapt the results presented in~\cite{DBLP:journals/tcs/DavidLMR15} to give some average-case properties about implicational bases.
We consider the base of proper premises and the Duquenne-Guigues base. 
We give the average size of the base of proper premises and show that the size of the base of proper premises is, on average, quasi-polynomial.
This implies that the size of the Duquenne-Guigues base is on average at most quasi-polynomial.
We then give an almost sure lower bound for the number of proper premises.

\medskip

The paper is organised as follows: in section~\ref{sec:defs} we introduce the definitions and notations that we use in the remainder of the paper.
Section~\ref{sec:average} contains the main results of this work.
In section~\ref{sec:discussion}, we discuss randomly generated contexts and the models that are used in this paper.
We then conclude and discuss future works.


\section{Definitions and Notations}\label{sec:defs}

In this section, we provide the definitions and results that will be used in this paper.
Most of the FCA definitions can be found in~\cite{DBLP:books/daglib/0095956}.
From now on, we will omit the brackets in the notation for sets when no confusion is induced by this simplification.

\subsection{Formal Concept Analysis}

A \emph{formal context} is a triple $\mathcal{C}=(\mathcal{O}, \mathcal{A}, \mathcal{R})$ in which $\mathcal{O}$ and $\mathcal{A}$ are sets of objects and attributes and $\mathcal{R}\subseteq \mathcal{O}\times \mathcal{A}$ is a binary relation between them.
A pair $(o,a)\in \mathcal{R}$  is read ``object $o$ has attribute $a$''.
Formal contexts can naturally be represented by cross tables, where a cross in the cell $(o,a)$ means that $(o,a)\in\mathcal{R}$.

\begin{table}
\centering
\begin{tabular}{c|ccccc}
 & $a_1$ & $a_2$ & $a_3$& $a_4$ & $a_5$\\
\hline
$o_1$ & $\times$& $\times$& & & \\

$o_2$ & & $\times$& &$\times$&$\times$\\

$o_3$ & &$\times$ &$\times$ &$\times$&\\

$o_4$ & & &$\times$& &$\times$\\

$o_5$ & & & &$\times$ &$\times$\\

\end{tabular}
\caption{Toy context $\mathcal{C}$.\label{ex:1}}
\end{table}

Table~\ref{ex:1} shows a toy context with 5 objects and 5 attributes.
It will serve as a running example throughout this paper.

\medskip

Let $O$ be a set of objects and $A$ a set of attributes, we denote by $O^\prime$ the set of all attributes that are shared by all objects of $O$ and $A^\prime$ the set of all objects that have all the attributes of $A$.
More formally, $O^\prime = \{a\in \mathcal{A}\ |\ \forall o\in O, (o,a)\in \mathcal{R}\}$ and $A^\prime = \{o\in \mathcal{O}\ |\ \forall a\in A, (o,a)\in \mathcal{R}\}$.

\medskip

The composition of those two operators, denoted $\cdot^{\prime\prime}$, forms a closure operator.
A set $X=X^{\prime\prime}$ is said to be closed.
A pair $(O,A)$ with $O\subseteq\mathcal{O}$, $A\subseteq\mathcal{A}$, $A^\prime = O$ and $O^\prime = A$ is called a \emph{(formal) concept} of the (formal) context $\mathcal{C}$.
In this case, we also have that $A^{\prime\prime} = A$ and $O^{\prime\prime} = O$. 

\medskip

The set of all the concepts of a context, ordered by inclusion on either their sets of attributes or objects forms a complete lattice.
Additionally, every complete lattice is isomorphic to the one formed by the concepts of a particular context.

\begin{definition}
An \emph{implication} (between attributes) is a pair of sets $X,Y\subseteq \mathcal{A}$.
It is noted $X\rightarrow Y$.
\end{definition}

\begin{definition}
An implication $X\rightarrow Y$ is said to \emph{hold} in a context $\mathcal{C}$ if and only if $X^\prime\subseteq Y^\prime$.
\end{definition}

\medskip
Many implications are redundant, that is if an implication $a\rightarrow c$ holds, then $ab\rightarrow c$ holds and is redundant.
The number of implications that hold can be quite large~\cite{DBLP:journals/jucs/Kuznetsov04}.
It is necessary to focus on the interesting ones.

\medskip

\begin{definition}
An implication set that allows for the derivation of all implications that hold in a context, and only them, through the application of Armstrong's axioms is called an implication base of the context.
\end{definition}

\begin{definition}[Duquenne-Guigues Base]
An attribute set $P$ is a pseudo-intent if and only if $P\not=P^{\prime\prime}$ and $Q^{\prime\prime}\subset P$ for every pseudo-intent $Q\subset P$.
The set of all the implications $P\rightarrow P^{\prime\prime}$ in which $P$ is a pseudo-intent is called the Duquenne-Guigues Base.
\end{definition}

The Duquenne-Guigues Base, also called \emph{canonical} base, or \emph{stem} base has first been introduced in~\cite{Guigues1986} and is the smallest (cardinality-wise) of all the bases.
Here, we denote this base as $\Sigma_{stem}$. The complexity of enumerating the elements of this base is studied in~\cite{DBLP:journals/dam/DistelS11}.

\subsubsection{Base of Proper Premises}

While the Duquenne-Guigues Base is the smallest base, the \emph{base of proper premises}, or \emph{Canonical Direct Base}, noted here $\Sigma_{Proper}$, is the smallest base for which the logical closure can be computed with a single pass.
The Canonical Direct Base was initially known under five independent definitions, shown to be equivalent by Bertet and Montjardet in~\cite{DBLP:journals/tcs/BertetM10}.

\medskip

\subsection{Hypergraphs and Transversals}

Let $V$ be a set of vertices.
A hypergraph $\mathcal{H}$ is a subset of the powerset $2^V$.
Each $E\in\mathcal{H}$ is called an (hyper)edge of the hypergraph. A set $S\subseteq V$ is called a hypergraph transversal of $\mathcal{H}$ if it intersects every edge of $\mathcal{H}$, that is $S\cap E\not=\emptyset, \forall E\in \mathcal{H}$.
A set $S\subseteq V$ is called a minimal hypergraph transversal of $\mathcal{H}$ if $S$ is a transversal of $\mathcal{H}$ and $S$ is minimal with respect to the subset inclusion among all the hypergraph transversals of $\mathcal{H}$.
The set of all minimal hypergraph transversals of $\mathcal{H}$ forms a hypergraph, that we denote $Tr(\mathcal{H})$ and that is called the transversal hypergraph.

\subsection{Proper Premises as Hypergraph Transversals}

In this section, we introduce a definition of the base of proper premises based on hypergraph transversals.

\begin{proposition}[from~\cite{DBLP:books/daglib/0095956}]
$P\subseteq \mathcal{A}$ is a premise of $a \in \mathcal{A}$ if and only if $(\mathcal{A}\setminus o^\prime)\cap P \not= \emptyset$ holds for all $o\in \mathcal{O}$ such that $(o,a)\not\in\mathcal{R}$.
$P$ is a proper premise for $m$ if and only if $P$ is minimal with respect to subset inclusion for this property.
\end{proposition}

Proposition 23 from~\cite{DBLP:books/daglib/0095956} uses $o\swarrow a$ instead of $(o,a)\not\in\mathcal{R}$.
It is a stronger condition that implies a maximality condition that is not necessary here.

\medskip

The set of proper premises of an attribute is equivalent to the minimal transversals of a hypergraph induced from the context with the following proposition:

\begin{proposition}[From~\cite{DBLP:journals/amai/RysselDB14}]
$P$ is a premise of $a$ if and only if $P$ is a hypergraph transversal of $\mathcal{H}_a$ where
\[\mathcal{H}_a = \{\mathcal{A}\setminus o^\prime|o\in \mathcal{O}, (o,a)\not\in\mathcal{R}\}\]
The set of all proper premises of $a$ is exactly the transversal hypergraph $Tr(\mathcal{H}_a)$.
\end{proposition}

To illustrate this link, we show the computation of the base of proper premises for context~\ref{ex:1}.
We must compute the hypergraph $\mathcal{H}_a$ for each attribute.
Let's begin with attribute $a_1$. We have to compute $\mathcal{H}_{a_1}=\{\mathcal{A}\setminus o^\prime~|o\in \mathcal{O}, (o,a_1)\not\in\mathcal{R}\}$ and $Tr(\mathcal{H}_{a_1})$. In $\mathcal{C}$, there is no cross for $a_1$ in the rows $o_2$, $o_3$, $o_4$ and $o_5$. We have :
\[\mathcal{H}_{a_1}=\{\{a_1,a_3\}, \{a_1,a_5\}, \{a_1,a_2,a_3\},\{a_1,a_2,a_4\}\}\] and \[Tr(\mathcal{H}_{a_1})=\{\{a_1\},\{a_2,a_3,a_5\},\{a_3,a_4,a_5\}\]

We have the premises for $a_1$, which give implications $a_2a_3a_5\rightarrow a_1$ and $a_3a_4a_5\rightarrow a_1$. 
$\{a_1\}$ is also a transversal of $\mathcal{H}_{a_1}$ but can be omitted here, since $a\rightarrow a$ is always true.

In the same way, we compute the hypergraph and its transversal hypergraph for all other attributes.
For example, 

\[\mathcal{H}_{a_2} = \{\{a_1,a_2,a_3\},\{a_1,a_2,a_4\}\}\mbox{ and } Tr(\mathcal{H}_{a_2})=\{\{a_1\},\{a_2\},\{a_3,a_4\}\}\]
\[\mathcal{H}_{a_5}=\{\{a_1,a_5\},\{a_3,a_4,a_5\}\}\mbox{ and } Tr(\mathcal{H}_{a_5}) = \{\{a_5\},\{a_1,a_3\},\{a_1,a_4\}\}\]
The set of all proper premises of $a_i$ is exactly the transversal hypergraph $Tr(\mathcal{H}_{a_i})$, $\forall i\in\{1,\dots,5\}$,  to which we remove the trivial transversals ($a_i$ is always a transversal for $\mathcal{H}_{a_i}$).
The base of proper premises for context $\mathcal{C}$ is the union of the proper premises for each attributes:

\[\Sigma_{Proper}(\mathcal{C})=\bigcup_{a\in\mathcal{A}} Tr(\mathcal{H}_a)\setminus a\]

\section{Average size of an implication base}\label{sec:average}

In~\cite{DBLP:journals/amai/RysselDB14}, Distel and Borchmann provide some expected numbers for proper premises and concept intents.
Their approach, like the one in~\cite{DBLP:journals/tcs/DavidLMR15}, uses the Erd\"os-R\'enyi model~\cite{erdos1960evolution} to generate random hypergraphs.
However, in~\cite{DBLP:journals/amai/RysselDB14}, the probability for each vertex to appear in a hyperedge is a fixed $0.5$ (by definition of the model) whereas the approach presented in~\cite{DBLP:journals/tcs/DavidLMR15} consider this probability as a variable of the problem and is thus more general.

\medskip

\subsection{Single parameter model}
\label{spmavg}
In the following, we assume all sets to be finite, and that $|\mathcal{O}|$ is polynomial in $|\mathcal{A}|$.
We call $p$ the probability that an object $o$ has an attribute $a$.
An object having an attribute is independent from other attributes and objects.
We denote by $q=1-p$ the probability that $(o,a)\not\in\mathcal{R}$.
The average number of hyperedges of $\mathcal{H}_a, \forall a\in A$, is $m=|\mathcal{O}|\times q$.
Indeed, $\mathcal{H}_{a_i}$ has one hyperedge for each $(o,a_i)\not\in\mathcal{R}$.
The probability of an attribute appearing in a hyperedge of $\mathcal{H}_{a_i}$ is also $q$.

\medskip

We note $n$ the number of vertices of $\mathcal{H}_a$.
At most all attributes appear in $\mathcal{H}_a$, so $n\leq |\mathcal{A}|$

\begin{proposition}[Reformulated from~\cite{DBLP:journals/tcs/DavidLMR15}]
\label{prop:number}
In a random hypergraph with $m$ edges and $n$ vertices, with $m=\beta n^\alpha, \beta>0\mbox{ and }\alpha>0$ and a probability $p$ that a vertex appears in an edge, there exists a positive constant $c$ such that the average number of minimal transversals is
\[O\left(n^{d(\alpha)log_{\frac{1}{q}}m+c\ln\ln m}\right)\]
with $q=1-p$, $d(\alpha) = 1$ si $\alpha\leq1$ and $d(\alpha)=\frac{(\alpha+1)^2}{4\alpha}$ otherwise.
\end{proposition}

\medskip

Proposition~\ref{prop:number} bounds the average number of minimal transversals in a hypergraph where the number of edges is polynomial in the number of vertices.
In~\cite{DBLP:journals/tcs/DavidLMR15}, the authors also prove that this quantity is quasi-polynomial.

From Prop.~\ref{prop:number} we can deduce the following property for the number of proper premises for an attribute.

\medskip

\begin{proposition}
\label{prop:sizePP}
In a random context with $|\mathcal A|$ attributes, $|\mathcal O|$ objects and probability $p$ that $(o,a)\in\mathcal R$ , the number of proper premises for an attribute is on average:

\[O\left(|\mathcal{A}|^{\left(d(\alpha)log_{\frac{1}{p}}\left(|\mathcal{O}|\times q\right) + c\ln\ln \left(|\mathcal{O}|\times q\right))\right)}\right)\]

and is quasi-polynomial in the number of objects.
\end{proposition}

\medskip
Proposition~\ref{prop:sizePP} states that the number of proper premises of an attribute is on average quasi-polynomial in the number of objects in a context where the number of objects is polynomial in the number of attributes.

\medskip

As attributes can share proper premises, $|\Sigma_{Proper}|$ is on average less than $|\mathcal A|\times O\left(|\mathcal A|^{(d(\alpha)log_{\frac{1}{p}}(|\mathcal O|\times q) + c\ln\ln (|\mathcal O|\times q))}\right)$

\medskip

Since $|\Sigma_{stem}| \leq |\Sigma_{Proper}|$, Prop.~\ref{prop:sizePP} immediately yields the following corollary:

\medskip
\begin{corollary}\label{coro:DG}
The average number of pseudo-intents in a context where the number of objects is polynomial in the number of attributes is less than or equal to:
\[|\mathcal{A}|\times O\left(|\mathcal{A}|^{\left(d(\alpha)log_{\frac{1}{p}}\left(|\mathcal{O}|\times q\right) + c\ln\ln \left(|\mathcal{O}|\times q\right))\right)}\right)\]
\end{corollary}
\medskip

Corollary~\ref{coro:DG} states that in a context where the number of object polynomial in the number of attributes, the number of pseudo-intents is on average at most quasi-polynomial.

\subsection{Almost sure lower bound for the size of the number of proper premises}
\label{spmlb}

An almost sure lower bound is a bound that is true with probability close to 1.
In~\cite{DBLP:journals/tcs/DavidLMR15}, the authors give an almost sure lower bound for the number of minimal transversals.

\begin{proposition}[Reformulated from~\cite{DBLP:journals/tcs/DavidLMR15}]
\label{prop:ASLBMT}In a random hypergraph with $m$ edges and $n$ vertices, and a probability $p$ that a vertex appears in an edge, the number of minimal transversals is almost surely greater than
\[\mathcal{L}_{MT} = n^{log_{\frac{1}{q}}m+O(\ln\ln m)}\]
\end{proposition}

Proposition~\ref{prop:ASLBMT} states the in a random context with probability $p$ that a given object has a given attribute, one can expect at least $\mathcal{L}_{MT}$ proper premises for each attribute.

\begin{proposition}~\label{prop:ASLB}
In a random context with $|\mathcal A|$ attributes, $|\mathcal O|$ objects and probability $q$ that a couple $(o,a)\not\in\mathcal R$, the size of $\Sigma_{Proper}$ is almost surely greater than
\[|\mathcal{A}|\times|\mathcal{A}|^{\left(log_{\frac{1}{p}}\left(|\mathcal{O}|\times q\right) + O(\ln\ln \left(|\mathcal{O}|\times q\right))\right)}\]
\end{proposition}

\medskip
As Prop~\ref{prop:ASLB} states a lower bound on the number of proper premises, no bound on the number of pseudo-intents can be obtained this way.

\subsection{Multi-parametric model}

In this section we consider a multi-parametric model.
This model is more accurate with respect to real life data.
In this model, each attribute $j$ has a probability $p_j$ of having an object.
All the attributes are not equiprobable.

We consider a context with $m$ objects and $n$ attributes.
The set of attributes is partitioned into 3 subsets:

\begin{itemize}
\item The set $U$ represents the attributes that appear in a lot of objects' descriptions (ubiquitous attributes). For all attributes $u\in U$ we have $q_u=1-p_u<\frac{x}{m}$ with $x$ a fixed constant.
\item The set $R$ represents rare events, that is attributes that rarely appear. For all attributes $r\in R$ we have that $p_r=1-\frac{1}{\ln n}$ tends to 0.
\item The set $F=\mathcal A\setminus (U\cup R)$ of other attributes.
\end{itemize}

\begin{proposition}[Reformulated from theorem 3~\cite{DBLP:journals/tcs/DavidLMR15}]
\label{mpmavg}
In the multi-parametric model, we have:
\begin{itemize}
\item If $|O\cup R| = O(\ln |\mathcal A|)$, then the number of minimal transversal is on average at most polynomial.
\item If $|R|= O{((\ln |\mathcal A|)^c)}$, then the number of minimal transversal is on average at most quasi-polynomial.
\item If $|R|=\Theta(|\mathcal A|)$, then the number of minimal transversal is on average at most exponential on $|R|$.
\end{itemize}
\end{proposition}

\medskip
Proposition~\ref{mpmavg} states that when most of the attributes are common (that is, are in the set $U$), $|\Sigma_{Proper}|$ is on average at most polynomial. 
When there is a logarithmic quantity of rare attributes (attributes in $R$), $|\Sigma_{Proper}|$ is  on average at most quasi-polynomial (in the number of objects). 
When most of the attributes are rare events, $|\Sigma_{Proper}|$ is on average at most exponential.

\medskip

As in the single parameter model, Prop.~\ref{mpmavg} also yields the same bounds on the number of pseudo-intents.

\section{Discussion on randomly generated contexts}\label{sec:discussion}

The topic of randomly generated contexts is important in FCA, most notably when used to compare performances of algorithms.
Since~\cite{DBLP:journals/jetai/KuznetsovO02}, a few experimental studies have been made.
In~\cite{DBLP:conf/cla/BorchmannH16}, the authors investigate the Stegosaurus phenomenon that arises when generating random contexts, where the number of pseudo-intents is correlated with the number of concepts~\cite{borchmann2011decomposing}.

\medskip

As an answer to the Stegosaurus phenomenon raised by experiments on random contexts, in~\cite{DBLP:conf/cla/Ganter11}, the author discusses how to randomly and uniformly generate closure systems on 7 elements.

\medskip

In~\cite{DBLP:conf/smc/RimsaSZ13}, the authors introduce a tool to generate less biased random contexts, avoiding repetition while maintaining a given density, for any number of elements. However this tool doesn't ensure uniformity.

\medskip	

The partition of attributes induced by the multi-parametric model allows for a structure that is close to the structure of real life datasets~\cite{DBLP:journals/tcs/DavidLMR15}.
However, we can't conclude theoretically on whether this model avoids the stegosaurus phenomenon discussed in~\cite{DBLP:conf/cla/BorchmannH16}. This issue would be worth further theoretical and experimental investigation.

\section{Conclusion}
In this paper, we used results on average-case combinatorics on hypergraphs to bound the average size of the base of proper premises.
Those results concerns only the proper premises, and can't be applied on the average number of pseudo-intents.
However, as the Duquenne-Guigues base is, by definition, smaller than the base of proper premises, the average size of the base of proper premises can serve as an average bound for the number of pseudo-intents.

This approach does not give indications on the number of concepts.
However, there exists some works on this subject~\cite{DBLP:conf/icdm/LhoteRS05, DBLP:journals/dam/EmilionL09}.

As the average number of concepts is known~\cite{DBLP:conf/icdm/LhoteRS05, DBLP:journals/dam/EmilionL09}, and this paper gives some insight on the average size of some implicational bases, future works can be focused on the average number of pseudo-intents.
It would also be interesting to study the average number of $n$-dimensional concepts or implications, with $n\geq 3$~\cite{DBLP:conf/iccs/LehmannW95, DBLP:journals/order/Voutsadakis02}.

\section*{Acknowledgments}
This research was partially supported by the European Union's ``\emph{Fonds Europ\'een de D\'eveloppement R\'egional (FEDER)}'' program.

\bibliographystyle{unsrt}
\bibliography{Biblio}

\end{document}